\documentclass[runningheads]{llncs}

\usepackage{eccv}
\usepackage{eccvabbrv}
\usepackage{graphicx}
\usepackage{booktabs}
\usepackage{amsmath}
\usepackage{multirow}
\usepackage{multicol}
\usepackage{indentfirst}
\usepackage{xcolor,colortbl}
\usepackage[accsupp]{axessibility}
\usepackage[pagebackref,breaklinks,colorlinks,citecolor=eccvblue]{hyperref}

\definecolor{ashgrey}{rgb}{0.7, 0.75, 0.71}
\definecolor{skyblue}{rgb}{0.53, 0.81, 0.92}
\definecolor{coralpink}{rgb}{0.97, 0.51, 0.47}
\definecolor{mossgreen}{rgb}{0.68, 0.87, 0.68}

\begin{document}

\title{OurDB: Ouroboric Domain Bridging for Multi-Target Domain Adaptive Semantic Segmentation} 

\titlerunning{OurDB}

\author{Seungbeom Woo\inst{1} \and
Geonwoo Baek\inst{1} \and
Taehoon Kim\inst{1} \and
Jaemin Na\inst{2} \and
Joong-won Hwang\inst{3} \and
Wonjun Hwang\inst{1}}

\authorrunning{S. Woo et al.}

\institute{Ajou University \and
Tech. Innovation Group, KT \and
ETRI \\
\email{\{enpko47, bkw0622, th951113\}@ajou.ac.kr, jaemin.na@kt.com jwhwang@etri.re.kr, wjhwang@ajou.ac.kr}}

\maketitle

\begin{abstract}  
  Multi-target domain adaptation (MTDA) for semantic segmentation poses a significant challenge, as it involves multiple target domains with varying distributions.
  The goal of MTDA is to minimize the domain discrepancies among a single source and multi-target domains, aiming to train a single model that excels across all target domains.
  Previous MTDA approaches typically employ multiple teacher architectures, where each teacher specializes in one target domain to simplify the task. However, these architectures hinder the student model from fully assimilating comprehensive knowledge from all target-specific teachers and escalate training costs with increasing target domains.
  In this paper, we propose an ouroboric domain bridging (OurDB) framework, 
  offering an efficient solution to the MTDA problem using a single teacher architecture. 
  This framework dynamically cycles through multiple target domains, aligning each domain individually to restrain the biased alignment problem, and utilizes Fisher information to minimize the forgetting of knowledge from previous target domains. We also propose a context-guided class-wise mixup (CGMix) that leverages contextual information tailored to diverse target contexts in MTDA. Experimental evaluations conducted on four urban driving datasets (i.e., GTA5, Cityscapes, IDD, and Mapillary) demonstrate the superiority of our method over existing state-of-the-art approaches.

\end{abstract}

\section{Introduction}
\label{sec:intro}

Semantic segmentation, a fundamental task in computer vision, entails categorizing each pixel in an image with specific class labels. Recent advancements in deep learning-based semantic segmentation approaches~\cite{FCN, PSPNet, DeepLab, SegFormer}, coupled with extensive datasets, have propelled significant progress.
However, obtaining pixel-level annotated data for semantic segmentation is not only costly but also time-consuming, prompting many studies to train their models on synthetic datasets~\cite{GTA5, Synthia}. Nevertheless, deploying these models in real-world scenarios often leads to performance degradation due to domain shift, which denotes disparities in data distributions between synthetic and real-world data.
To alleviate this issue, unsupervised domain adaptation (UDA) methods~\cite{FCNDA, CyCADA, BDL, ADVENT, DCAN} have been proposed to mitigate the domain discrepancy between the source (e.g., \textit{synthetic}) and target (e.g, \textit{real-world}) domains.

\begin{figure}[t]
    \centering
    \includegraphics[width=0.8\linewidth]{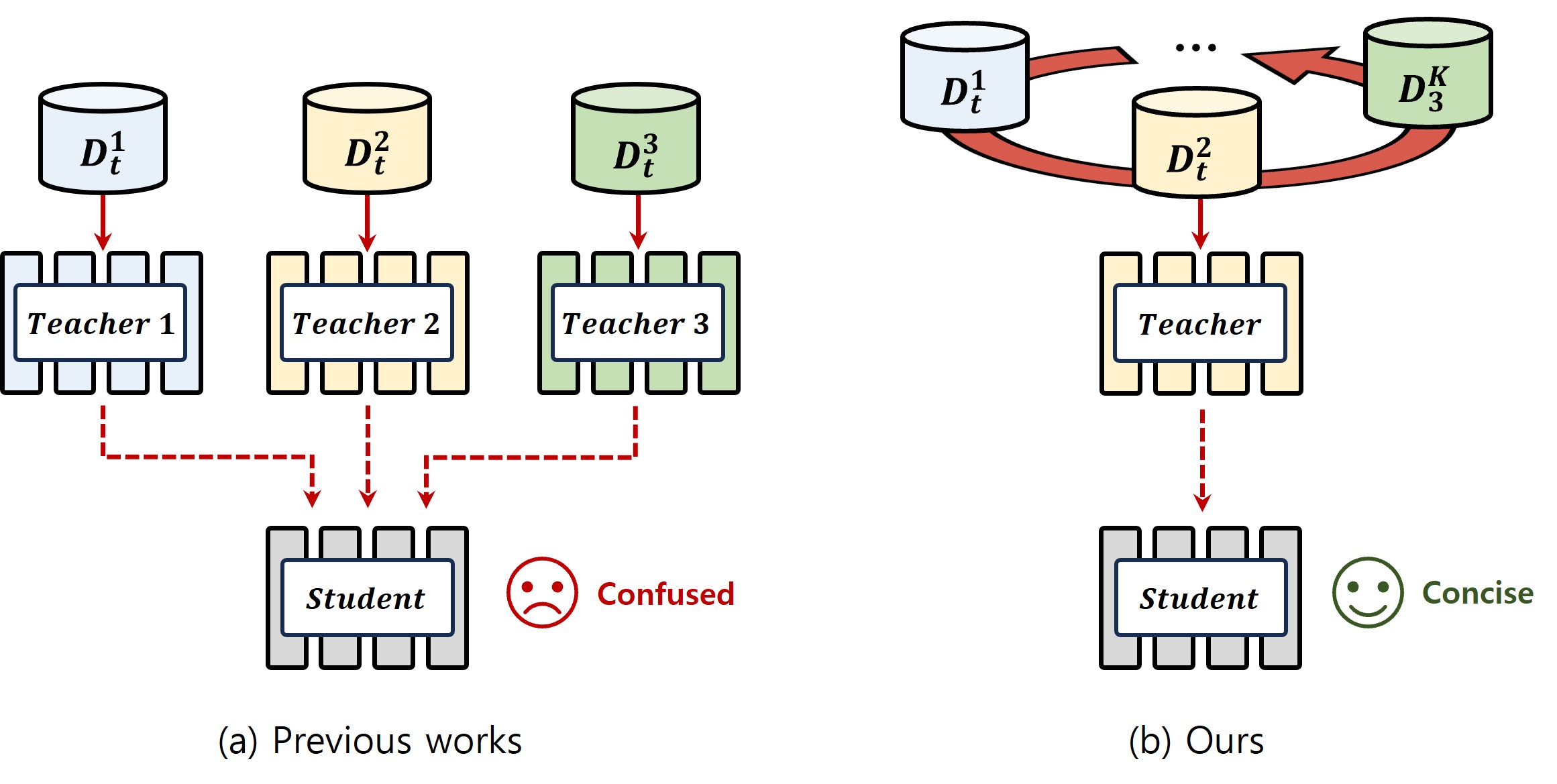}
    \vspace{-3mm}
    \caption{Concept comparison between previous works and ours for MTDA. (a) Previous methods adopt a multiple teacher architecture to address the multiple target adaptation challenge. (b) In contrast, ours employs a single teacher architecture, leveraging cyclic adaptation across multiple domains akin to the ouroboros, a symbol of cyclic renewal and self-sustainment. }
    \label{fig:figure1}
    \vspace{-5mm}
\end{figure}

Traditional UDA methods typically work under a \textit{single-source-single-target} setting, assuming that the source and target data are drawn from a distinct single underlying distribution. However, this assumption overlooks the multi-domain nature of real-world datasets. For instance, in autonomous driving systems, test data is often collected across various lighting conditions, weather patterns, and urban landscapes.
There are two naive approaches to extend traditional UDA methods to multiple target domains: \textit{Data Combination} and \textit{Individual Model}. \textit{Data Combination} treats multiple target domains as a unified domain for adaptation, while \textit{Individual Model} entails training separate models for each target domain. However, \textit{Data Combination} disregards domain shifts among the target domains, resulting in biased alignment~\cite{DHA}, where only the target domain most similar to the source domain is excessively adapted. Conversely, \textit{Individual Model} increases memory complexity significantly due to the operation of multiple models. To overcome these limitations, the concept of Multi-Target Domain Adaptation (MTDA) has been introduced.

MTDA comprises a single labeled source domain and multiple unlabeled target domains, where the multiple target domains have their own different distributions. The goal of MTDA is to develop a single model capable of achieving high performance across multiple target domains. 
In previous methods~\cite{CCL, MTKT, CoaST} for semantic segmentation, multiple teacher architectures are commonly employed to achieve this objective. In such architectures, each teacher model concentrates on one target domain for adaptation and imparts its knowledge to a domain-agnostic student model. While employing multiple teacher architectures offers the advantage of breaking down a \textit{single-source-multi-target} task into a kind of \textit{single-source-single-target} tasks, thereby providing an avenue to start from traditional UDA methods. However, these architectures endow each teacher model with unique domain knowledge. Consequently, as illustrated in~\cref{fig:figure1} (a), when a student model is simultaneously distilled knowledge from disparate teacher models, it may face difficulty in fully integrating comprehensive knowledge about all target domains, potentially leading to sub-optimal performance~\cite{CCL}.

In this paper, we propose an ouroboric domain bridging (OurDB) framework, which employs a single teacher architecture based on the mean-teacher method~\cite{MeanTeacher}. 
To address the biased alignment that arises from adapting all target domains with a single teacher model, we introduce an ouroboric domain selector (ODS) module, as shown in~\cref{fig:figure1} (b). This module cyclically engages with multiple target domains, focusing on one domain at a time during the learning process. Essentially, the ODS module ensures that a teacher model aligns efficiently with only one source-target domain pair in each iteration. However, as we cycle through domains and teach the model one domain at a time, there is a risk of diluting knowledge from the previous domain. To counteract this issue, OurDB incorporates an anti-forgetting EMA (AF-EMA) mechanism, which leverages Fisher information~\cite{EWC} to prevent the teacher model from forgetting the knowledge acquired from previous domains. Furthermore, we introduce context-guided class-wise mixup (CGMix), which builds upon ClassMix~\cite{ClassMix}, adapting it to perform domain adaptation across multiple target domains that may exhibit diverse contexts.
The main contributions of this paper are summarized as follows:
\begin{itemize}
    \item We propose OurDB framework, comprising ODS, AF-EMA, and CGMix to efficiently address the MTDA problem using a single teacher architecture. 
    \item ODS module cyclically engages with multiple target domains, concentrating on one domain per iteration to ensure efficient alignment with source-target domain pairs, and AF-EMA prevents the teacher model from forgetting knowledge acquired from the previous domain.
    \item We present CGMix tailored for domain adaptation across multiple target domains with diverse contexts. 
    \item We conduct experiments on several MTDA protocols using GTA5, Cityscapes, IDD, and Mapillary to demonstrate the superiority of our method.
\end{itemize}

\section{Related Works}

\textbf{Unsupervised Domain Adaptation for Semantic Segmentation.} UDA is a task that has recently garnered significant attention. The goal of UDA is to learn a model by transferring knowledge from a labeled source domain to an unlabeled target domain under the domain shift. This advantage has prompted many studies on UDA in semantic segmentation tasks, where the collection of labeled data is expensive. To reduce the domain discrepancy between the source and target domains, the methods of UDA for semantic segmentation can be broadly grouped into three categories. The first category is adversarial learning-based approaches~\cite{FCNDA, CLAN, ADVENT, JAL}, which aim to learn a domain-invariant representation through a min-max game between a feature extractor and a domain discriminator. The second one involves self-training approaches~\cite{BDL, ESL, Uncertainty, ProDA}, which generate pseudo labels for the target data and subsequently learn from these pseudo labels to obtain knowledge of the target domain. Some works~\cite{BDL, ESL} generate reliable pseudo labels by thresholding based on confidence or entropy, while others~\cite{Uncertainty, ProDA} weight pseudo labels through class centroid or uncertainty. Lastly, domain bridging approaches~\cite{CyCADA, DCAN, DACS, DDB, BDM} form the third category. Instead of directly adapting from the source domain to the target domain, domain bridging methods build bridges between the source and target domains to adapt gradually. Some studies~\cite{CyCADA, DCAN} utilize source data, which is made to resemble the target domain through style transfer, as a bridge, while others~\cite{DACS, DDB, BDM} take advantage of bridges created by mix-based data augmentation. Although UDA for semantic segmentation has been extensively studied and has made significant progress, most works address the UDA problem under the single-source-single-target setting. This assumption limits the applicability of traditional UDA methods to real-world applications with multiple domains. Therefore, it is necessary to study the MTDA, which is more suitable for the real world.

\textbf{Multi-Target Domain Adaptation for Semantic Segmentation.} The MTDA for semantic segmentation has not been actively explored, as it is a more challenging task compared to UDA. In the MTDA problem, domain discrepancies are present not only between the source and target domains, but also among the target domains. Several studies~\cite{CCL, MTKT, ADAS, CoaST} have been proposed to address this problem. Isobe \textit{et al.}~\cite{CCL} are the first to propose a method for MTDA in the semantic segmentation task. They employ multiple teacher architectures to effectively align multiple target domains. Each teacher focuses solely on one target domain and aligns a source-target domain pair through style transfer. The knowledge from all teacher models is then transferred to the domain-generic student model using knowledge distillation (KD)~\cite{HintonKD}. Saporta \textit{et al.} propose two adversarial frameworks to tackle the MTDA problem: multi-discriminator and multi-target knowledge transfer (MTKT). The multi-discriminator uses two discriminators to perform source-target alignment and target-target alignment. MTKT aligns each source-target domain pair through multiple target-specific classifiers and transfers the knowledge from all target-specific classifiers to a target-agnostic classifier. Subsequently, ADAS~\cite{ADAS} and CoaST~\cite{CoaST} are proposed, which both utilize style transfer for adaptation. ADAS introduces a new statistics-based style transfer method that can transform source data into diverse target styles. CoaST, as with MTKT, utilizes multiple domain-specific classifiers for MTDA. Each of these domain-specific classifiers is trained on style-transferred source data through self-training and distills its knowledge to a domain-agnostic classifier.

Compared with the previous methods~\cite{CCL, MTKT, CoaST}, we present a novel approach that effectively addresses MTDA with only a single teacher architecture. 
To attain this objective, we introduce an ouroboric domain bridging framework that cyclically adapts by selecting one domain from multiple target domains.
Furthermore, we utilize a novel augmentation method started from ClassMix~\cite{ClassMix} instead of style transfer to minimize the domain gaps among multiple domains.

\section{Proposed Method}

\begin{figure}[tb]
  \centering
  \includegraphics[width=1.0\textwidth]{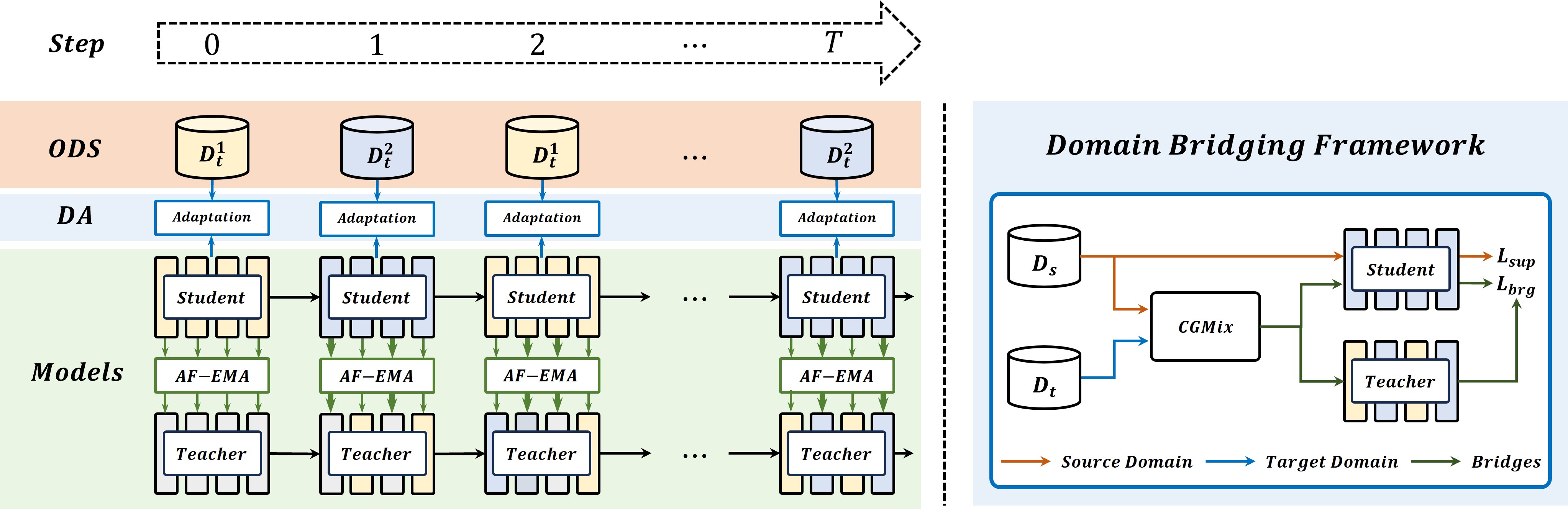}
  \caption{Overview of the proposed OurDB. The framework is illustrated as an example where the number of target domains is $K=2$. ODS cyclically selects one domain to be aligned from multiple target domains (\textit{\textcolor{coralpink}{red area}}). OurDB then learns the bridges generated by CGMix to gradually reduce the domain discrepancies between the source and selected target domains (\textit{\textcolor{skyblue}{blue area}}). AF-EMA adaptively adjusts the EMA coefficients based on Fisher information to prevent the teacher model from forgetting the knowledge about the previous target domain (\textit{\textcolor{mossgreen}{green area}}).}
  \label{fig:overview}
  \vspace{-3mm}
\end{figure}

OurDB introduces three methods to solve MTDA based on only a single teacher architecture: \lowercase\expandafter{\romannumeral1}) ODS, \lowercase\expandafter{\romannumeral2}) AF-EMA, and \lowercase\expandafter{\romannumeral3}) CGMix. The overall framework is illustrated in \cref{fig:overview}.

\subsection{Preliminaries}

\textbf{Problem definition.} Our main goal is to train a model $f_{\theta}:\mathbb{R}^{H \times W \times 3} \rightarrow \mathbb{R}^{H \times W \times C}$ that performs effectively across all target domains, where $H$ and $W$ denote the height and width of the images, respectively, and $C$ represents the number of categories. We assume that we have access to a single source domain $\mathcal{D}_s$ and $K$ multiple target domains $\mathcal{D}_t = \{\mathcal{D}_t^k\}_{k=1}^K$. The source domain $\mathcal{D}_s$ consists of $N_s$ images $x_s \in \mathbb{R}^{H \times W \times 3}$ with their corresponding ground-truth labels $y_s \in \mathbb{R}^{H \times W \times C}$. The $k$-th target domain $\mathcal{D}_t^k$ is composed of $N_t^k$ images $x_t^k \in \mathbb{R}^{H \times W \times 3}$ without corresponding ground-truth labels. As with the Standard MTDA setting, we suppose that the marginal distributions among all domains are different, while the label space is shared.

\textbf{Mean-teacher framework.} The mean-teacher framework~\cite{MeanTeacher}, commonly employed in UDA, has a teacher and a student model. The student model $f_{\theta}$ is trained on labeled source and unlabeled target data as follows:
\begin{align}
    \begin{split}
        \mathcal{L}_{sup} &= \mathcal{L}_{ce}\left(f_{\theta}(x_s), y_s\right), \\
        \mathcal{L}_{unsup} &= \mathcal{L}_{ce}\left(f_{\theta}(x_{t}), \hat{y}_{t}\right),
    \end{split}
\end{align}
where $\mathcal{L}_{sup}$ and $\mathcal{L}_{unsup}$ denote the supervised loss on the source data and the unsupervised loss on the target data, respectively, and $\mathcal{L}_{ce}$ represents a pixel-wise cross-entropy loss. $\hat{y}_t$ is the pseudo labels generated by the teacher model.

To enhance the stability of pseudo labels for target data, this framework constructs the teacher model $f_{\theta^\prime}$ as a temporal ensemble of the student model $f_{\theta}$ through an exponential moving average (EMA) as follows:
\begin{equation}
    \theta^\prime = \alpha\theta^\prime + (1 - \alpha)\theta,
\end{equation}
where $\alpha \in [0, 1]$ is an EMA coefficient, and $\theta^{\prime}$ and $\theta$ denote the parameters of the teacher and student models, respectively. In this work, we build the OurDB based on the mean teacher framework.

\textbf{ClassMix.} ClassMix~\cite{ClassMix} is a well-designed data augmentation technique that blends two data by respecting semantic boundaries in semantic segmentation. In UDA for semantic segmentation, ClassMix is used to build bridges $(x_{brg},\hat{y}_{brg})$ between source and target domains. Half of the classes from the source data are randomly selected and pasted into the target data as:
\begin{align}
    \begin{split}
        x_{brg} &= \mathbf{M}_{cls} \odot x_s + (\mathbf{1} - \mathbf{M}_{cls}) \odot x_t, \\
        \hat{y}_{brg} &= \mathbf{M}_{cls} \odot y_s + (\mathbf{1} - \mathbf{M}_{cls}) \odot \hat{y}_t,
    \end{split}
\end{align}
where $\mathbf{M}_{cls} \in \{0, 1\}^{H \times W}$ represents class masks, with pixels belonging to the selected classes are set to 1 and the others are set to 0. $\odot$ is the element-wise multiplication operator. The model $f_{\theta}$ then trains the bridges to align the source and target domains as follows:
\begin{equation}
   \mathcal{L}_{brg} = \mathcal{L}_{ce}\left(f_{\theta}(x_{brg}), \hat{y}_{brg}\right),
\end{equation}
where $\mathcal{L}_{brg}$ denotes the bridging loss. OurDB leverages a ClassMix-based domain bridging technique to reduce the domain gap between source and target domains.

\subsection{Ouroboric Domain Bridging for MTDA}

\textbf{Ouroboric domain selector.} From a human student's perspective, it would be difficult to understand when multiple subject teachers are simultaneously instructing different subjects to a single student. Similarly, in MTDA, the biased alignment~\cite{DHA} occurs when simultaneously aligning multiple target domains that have different characteristics with a single teacher architecture. To resolve this issue, we need to convert MTDA to the \textit{single-source-single-target} task. We propose an ODS to accomplish this purpose. ODS is a module that cyclically selects one domain from among multiple target domains for adaptation (e.g., $target1 \rightarrow target2 \rightarrow \cdots \rightarrow targetK \rightarrow target1 \rightarrow \cdots$) as shown in~\cref{fig:figure1} (b). With the help of this module, OurDB concentrates on aligning only one source-target domain pair in every iteration.

Since ODS cyclically switches the target domain used for training, defining the switching interval is crucial. The switching interval can be based on iteration or epoch. An iteration-based switching interval causes ODS to change target domains every pre-defined iteration step. However, this interval necessitates hyper-parameter tuning for the iteration step. If the switching interval is too short, the target domain is replaced before the model has learned enough. Conversely, if the switching interval is too long, the model overfits to the current target domain and forgets the knowledge learned from the previous target domain. Therefore, in this work, we adopt an epoch-based switching interval to circumvent the need for such hyper-parameter tuning.

An epoch-based switching interval allows ODS to replace the target domain after the model has learned all the data in the current one. The formal definition of the epoch-based switching interval for ODS is as follows:
\begin{equation}
    \mathcal{D}_t^k = 
    \begin{cases}
        \mathcal{D}_t^{k+1} & \mbox{if \textit{epoch is completed}} \\
        \mathcal{D}_t^k & \mbox{otherwise}
    \end{cases}.
\end{equation}

Although an epoch-based switching interval ensures the model learns enough of the current target domain, it can lead to overfitting. Thus, we propose an anti-forgetting EMA to mitigate the teacher model from forgetting the knowledge of the previous target domain.

\textbf{Anti-forgetting EMA.} Let's come back to the human student perspective once again. Cycling through multiple subjects can lead the student to forget what he has learned before. The similar problem can arise in MTDA, and in this paper we try to find out a solution through continual learning methods~\cite{EWC, LwF, iCaRL, SSUL}. 
In continual learning, the methods to alleviate the forgetting of knowledge gained from previous tasks are being actively researched. Kirkpatrick \textit{et al.}~\cite{EWC} proposed a regularization term to conserve the knowledge derived from previous tasks while assimilating knowledge about the current task through Fisher information. Fisher information generally exhibits higher values for model's parameters that contain valuable information on previously learned tasks. Taking this observation, the regularization term based on Fisher information updates less the parameters that are important for the previous task. Inspired by~\cite{EWC}, we propose an AF-EMA to prevent the teacher model from forgetting the knowledge about the previous domain by adaptively adjusting the EMA coefficient based on Fisher information.

Upon the completion of the epoch for the current target domain $\mathcal{D}_t^k$, the Fisher information matrix $\mathcal{F}$ for teacher model's parameters $\theta^\prime$ is calculated as follows:
\begin{equation}
    \mathcal{F} = \frac{1}{N_t^k} \sum_{x_t^k \in \mathcal{D}_t^k} \left(\nabla_{\theta^\prime} \mathcal{L}_{ce}(f_{\theta^\prime}(x_t^k), \hat{y}_t^k) \right)^2.
\end{equation}

In order to utilize the Fisher information matrix as the EMA coefficients, every constituent of this matrix should have a value between 0 and 1 inclusive. We manipulate the Fisher information matrix through normalization and clipping as follows:
\begin{equation}
    \mathcal{F}^\prime = clip(norm(\mathcal{F}), \lambda_1, \lambda_2),
\end{equation}
where $clip(\cdot)$ and $norm(\cdot)$ represent the clipping function and the min-max normalization function, respectively. $[\lambda_1, \lambda_2]$ is the range of the EMA coefficients. The teacher model's parameters $\theta^\prime$ then are updated based on the adjusted Fisher information matrix $\mathcal{F}^\prime$ as follows:
\begin{equation}
    \theta^\prime = \mathcal{F}^\prime\theta^\prime + (1 - \mathcal{F}^\prime)\theta.
\end{equation}

\subsection{Context-Guided Class-wise Mixup}

Each domain inherently forms its own unique context. In MTDA, there exists a variety of target domains, thus in contrast to UDA, the model confronts diverse contexts. As the domain varies, the performance of the model may degrade due to mismatches in context. These context mismatches predominantly arise from disparate spatial distributions across different domains. In other words, the spatial locations where classes are primarily distributed can be varied across domains. ClassMix~\cite{ClassMix} pastes the source classes into the target data without spatial adjustments. Therefore, bridges generated by ClassMix can have distorted contextual information under the assumption of domain shift. The model trained with these bridges relies heavily on the characteristics of the classes and fails to make predictions that can be readily categorized by contextual information. Although the domain undergoes a shift, the contextual relationships among classes remain consistent. For example, cars are almost located on the road, not in the sky. We propose a CGMix for MTDA, which adjusts the spatial locations of classes based on these contextual relationships, thereby building bridges without contaminating contextual information.

\textbf{Context vectors for contextual relationships.} We measure contextual relationships through a histogram of neighboring pixels about the classes to paste into the target data. This histogram is designated as a \textit{context vector}. We create bridges with untainted contextual information by attaching the classes to the spatial location where the context vector is most similar in the target data. For this purpose, we generate a neighbor mask $\mathbf{M}_{nbr} \in \{0, 1\}^{H \times W}$ to extract the neighboring pixels of specific classes as follows:
\begin{equation}
    \mathbf{M}_{nbr} = \mathbb{I}\left[ \mathcal{G}(\mathbf{M}_{cls}) > 0\right] - \mathbf{M}_{cls},
\end{equation}
where $\mathbb{I}[\cdot]$ represents an indicator function, and $\mathcal{G}(\cdot)$ is the Gaussian filter. We then compute the context vector through neighbor mask as follows:
\begin{align}
    \begin{split}
        c_s &= hist(y_s \odot \mathbf{M}_{nbr}),
    \end{split}
\end{align}
where $hist(\cdot)$ denotes the histogram function, $c_s$ is a context vector in the source.

\textbf{CGMix.} The optimal spatial location for pasting classes in the target data is determined by inserting these classes into various locations in the target data and comparing the resulting context vectors. CGMix generates $N_{aug}$ augmented data $(X_{aug}, Y_{aug})$ in which selected classes exists in different locations by applying geometric data augmentation (e.g., horizontal flipping, translation, scaling) to the source data. We attach the classes from the augmented data into the target data, subsequently calculating the context vectors $\mathbb{C}_t$ as follows:
\begin{equation}
    \mathbb{C}_t = hist(\hat{y}_t \odot \mathbf{M}_{nbr}^\prime),
\end{equation}
where $\mathbf{M}_{nbr}^\prime$ is a neighbor mask for augmented data. CGMix then computes the cosine similarities between $c_s$ and $\mathbb{C}_t$, and selects the augmented sample $(x_{aug}, y_{aug})$ that exhibits the highest similarity to $c_s$ as follows:
\begin{align}
    \begin{split}
        x_{aug} &= X_{aug}[{\arg\max(cos(c_s, \mathbb{C}_t))}], \\
        y_{aug} &= Y_{aug}[{\arg\max(cos(c_s, \mathbb{C}_t))}],
    \end{split}
\end{align}
where $cos(\cdot)$ denotes the cosine similarity function. We build a bridge that has uncontaminated contextual information by leveraging this sample as follows:
\begin{align}
    \begin{split}
        x_{brg} &= \mathbf{M}_{cls}^\prime \odot x_{aug} + (\mathbf{1} - \mathbf{M}_{cls}^\prime) \odot x_t, \\
        \hat{y}_{brg} &= \mathbf{M}_{cls}^\prime \odot y_{aug} + (\mathbf{1} - \mathbf{M}_{cls}^\prime) \odot \hat{y}_t,
    \end{split}
\end{align}
where $\mathbf{M}_{cls}^\prime$ represents the class mask for an augmented sample.

\section{Experiments}

\subsection{Experimental setting}

\textbf{Datasets.} We conduct experiments on four urban driving datasets to demonstrate the superiority of OurDB. The training datasets include one synthetic dataset (i.e., GTA5~\cite{GTA5}) and three real-world datasets (i.e., Cityscapes~\cite{Cityscapes}, IDD~\cite{IDD}, and Mapillary~\cite{Mapillary}), each with different visual appearance.
\begin{itemize}
    \item \textbf{GTA5} is a synthetic dataset generated by the game engine and consists of 24,966 samples.
    \item \textbf{Cityscapes} is a real-world dataset from 50 European cities, primarily collected in Germany. It contains 2,975 training and 500 validation samples.
    \item \textbf{IDD} is a real-world dataset collected in India and comprises 6,993 training and 981 validation samples.
    \item \textbf{Mapillary} is a real-world dataset collected from various cities around the world, comprising 18,000 training and 2,000 validation samples.
\end{itemize}

\textbf{Experimental protocols.} Standard protocols for MTDA in semantic segmentation are typically constructed by combining the four mentioned urban driving datasets. These protocols are categorized into two types based on the label mapping method: (\lowercase\expandafter{\romannumeral1}) The 7 classes benchmark, as introduced in~\cite{MTKT}, which considers 7 super categories and requires the resolution of both training and evaluation images to be downscaled to 640 $\times$ 320. (\lowercase\expandafter{\romannumeral2}) The 19 classes benchmark, as introduced in~\cite{CCL}, which considers 19 categories and necessitates the resizing of images to a resolution of 1024 $\times$ 512 for training and evaluation. The model's performance is evaluated using the mIoU (mean Intersection over Union) metric, with the average mIoU across all target domains used as the final performance measure.

\textbf{Implementation details.} For a fair comparison, we conduct experiments under the same conditions as in previous works~\cite{CCL, MTKT, ADAS, CoaST}. We use DeepLab-V2~\cite{DeepLab} as the segmentation model and ResNet-101~\cite{ResNet} as the pre-trained backbone on ImageNet~\cite{ImageNet}. The segmentation model is trained for 120K iterations using a stochastic gradient descent optimizer with a learning rate of $2.5\times10^{-4}$. We set the momentum and weight decay to 0.9 and $5\times10^{-4}$, respectively. We also employ the polynomial learning rate scheduler~\cite{DeepLab}. In all experiments, the batch size is set to 4. For AF-EMA, the values of $\lambda_1$ and $\lambda_2$ are set to 0.99 and 0.9999, respectively. For CGMix, we assign a value of 10 to $N_{aug}$.

\subsection{Main Results}

\begin{figure*}[t]
    \centering
    \vspace{-4mm}
    \includegraphics[width=1\linewidth]{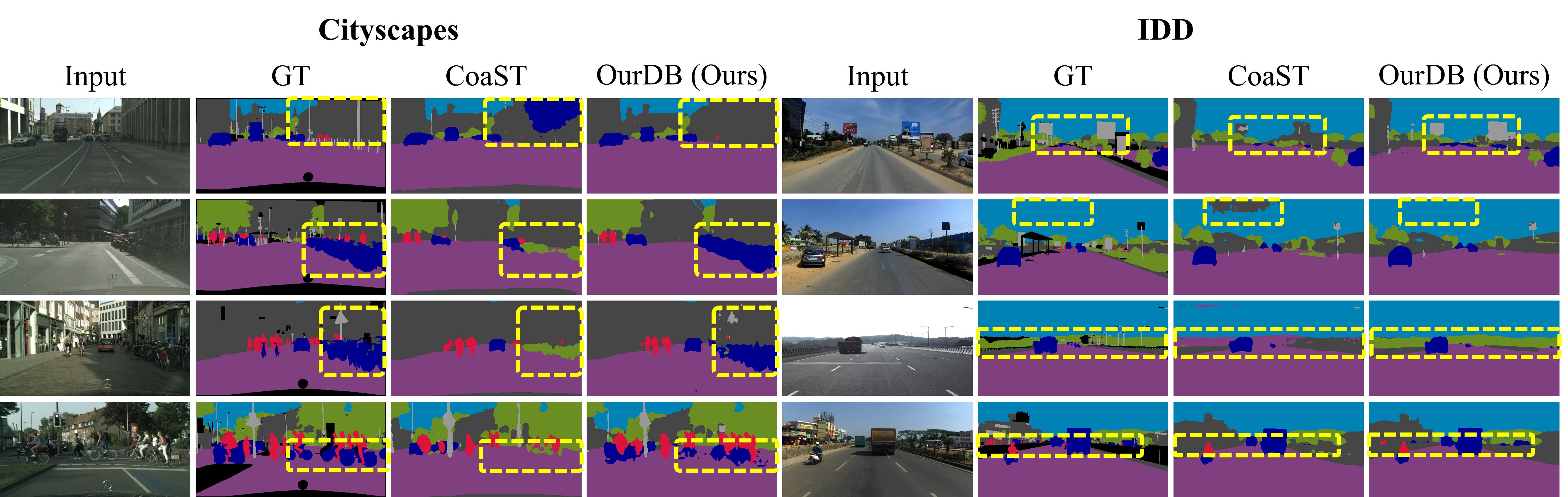}
    \vspace{-5mm}
    \caption{Qualitative comparison between the CoaST and OurDB on the 7 classes benchmark for the GTA5 $\rightarrow$ Cityscapes + IDD combination.}
    \label{fig:qualitative}
    \vspace{-2mm}
\end{figure*}

\textbf{7 classes benchmark.}~\cref{tab:cls7} presents results on the 7 classes benchmark for GTA5 $\rightarrow$ Cityscapes $+$ IDD combination. To demonstrate the effectiveness of our method on the MTDA problem, we consider both naive approaches extended from UDA methods (i.e., Data Combination, Individual Model) and the MTDA methods (i.e., Multi-Discriminator~\cite{MTKT}, MTKT~\cite{MTKT}, ADAS~\cite{ADAS}, CoaST~\cite{CoaST}). Data Combination and Individual Model are the approaches mentioned in~\cref{sec:intro}, and they are trained employing ADVENT~\cite{ADVENT} as a UDA method. OurDB achieves 73.5$\%$ performance in terms of average mIoU. This represents a performance improvement of +6.0$\%$ over the Individual Model, a UDA method, and a +2.2$\%$ gain over the CoaST, an MTDA method. Specifically, our method surpasses the CoaST by +3.2$\%$ on the Cityscapes dataset. On the more challenging IDD dataset, the performance gain is slightly lower at +1.2$\%$, yet OurDB exhibits robust performance on classes associated with small objects, i.e., \textit{object}, \textit{human}, and \textit{vehicle}. These results are further demonstrated in~\cref{fig:qualitative}, which provides a qualitative comparison. The comparison results for all combinations on the 7 classes benchmark are available in~\cref{tab:cls7_all}. We attain higher performance against the state-of-the-art methods in all experiments, including the scenario aligning three target domains. This proves that OurDB is effective across diverse target domains.

\begin{table*}[t]
\centering
\caption{Comparison results on the 7 classes benchmark for the GTA5 $\rightarrow$ Cityscapes $+$ IDD combination. The best score is indicated in bold and the second best one is underlined.}
\vspace{-2mm}
{\scriptsize 
\begin{tabular}{c|c|ccccccc|c|c}
\midrule[1px]
Method              & Target & flat & constr. &  object & nature & sky & human & vehicle & mIoU & Avg.         \\ \midrule[0.5pt]\midrule[0.5pt]
\multirow{2}{*}{Data Comb.~\cite{ADVENT}}   & C               & 93.9          & 80.2             & 26.2            & 79.0            & 80.5         & 52.5           & 78.0             & 70.0          & \multirow{2}{*}{67.4} \\
                              & I               & 91.8          & 54.5             & 14.4            & 76.8            & 90.3         & 47.5           & 78.3             & 64.8          &                       \\ \midrule[0.5pt]
\multirow{2}{*}{Individual~\cite{ADVENT}}   & C               & 93.5          & 80.5             & 26.0            & 78.5            & 78.5         & 55.1           & 76.4             & 69.8          & \multirow{2}{*}{67.5} \\ 
                              & I               & 91.2          & 53.1             & 16.0            & 78.2            & 90.7         & 47.9           & 78.9             & 65.1          &                       \\  \midrule[0.5pt]
\multirow{2}{*}{Multi-Dis.~\cite{MTKT}}   & C               & 94.3          & 80.7             & 20.9            & 79.3            & 82.6         & 48.5           & 76.2             & 68.9          & \multirow{2}{*}{67.3} \\
                              & I               & 92.3          & 55.0             & 12.2            & 77.7            & 92.4         & 51.0           & 80.2             & 65.7          &                       \\ \midrule[0.5pt]
\multirow{2}{*}{MTKT~\cite{MTKT}}         & C               & 94.5          & 82.0             & 23.7            & 80.1            & 84.0         & 51.0           & 77.6             & 70.4          & \multirow{2}{*}{68.2} \\  
                              & I               & 91.4          & 56.6             & 13.2            & 77.3            & 91.4         & 51.4           & 79.9             & 65.9          &                       \\  \midrule[0.5pt]
\multirow{2}{*}{ADAS~\cite{ADAS}}         & C               & \underline{95.1}          & 82.6             & \textbf{39.8}           & \textbf{84.6}           & 81.2         & \textbf{63.6}          & \underline{80.7}             & \underline{75.4}          & \multirow{2}{*}{71.2} \\ 
                              & I               & 90.5          & \textbf{63.0}            & \underline{22.2}            & 73.7            & 87.9         & 54.3           & 76.9             & 66.9          &                       \\  \midrule[0.5pt]
\multirow{2}{*}{CoaST~\cite{CoaST}}        & C               & 94.7          & \underline{82.9}             & 25.4            & 82.2            & \textbf{88.2}        & 54.4           & 80.5             & 72.6          & \multirow{2}{*}{71.3} \\ 
                              & I               & \textbf{94.2}          & \underline{61.5}             & 20.0            & \underline{82.7}            & \textbf{93.4}         & \underline{55.5}           & \underline{82.6}             & \underline{70.0}          &                       \\  \midrule[0.5pt] \midrule[0.5pt] \rowcolor{gray!20}

                              & C               & \textbf{96.1}             & \textbf{85.4}                 & \underline{34.9}                 & \underline{82.7}              & \underline{86.6}              & \underline{60.4}              & \textbf{84.4}                  & \textbf{75.8}              &      \\ \rowcolor{gray!20}
\multirow{-2}{*}{OurDB (Ours)}& I               & \underline{93.7}             & 60.0                & \textbf{29.6}                 & \textbf{83.0}                & \underline{92.7}             & \textbf{56.7}                & \textbf{82.9}                  & \textbf{71.2}               & \multirow{-2}{*}{\textbf{73.5}}                   \\  \midrule[1pt]
\end{tabular}}
\label{tab:cls7}
\end{table*}

\begin{table*}[t!]
\centering
\caption{Comparison results on the 19 classes benchmark for the GTA5 $\rightarrow$ Cityscapes $+$ IDD combination. The best score is indicated in bold and the second best one is underlined.}
\vspace{-2mm}
\resizebox{\textwidth}{!}{\tiny
\begin{tabular}{c|c|ccccccccccccccccccc|c|c}
\midrule[1px]
Method              & \rotatebox{90}{Target} & \rotatebox{90}{road} & \rotatebox{90}{sidewalk} & \rotatebox{90}{building} & \rotatebox{90}{walk}& \rotatebox{90}{fence} & \rotatebox{90}{pole} & \rotatebox{90}{light} & \rotatebox{90}{sign} & \rotatebox{90}{veg.} & \rotatebox{90}{terrain} & \rotatebox{90}{sky} & \rotatebox{90}{person} & \rotatebox{90}{rider} & \rotatebox{90}{car} & \rotatebox{90}{truck} & \rotatebox{90}{bus} & \rotatebox{90}{train} & \rotatebox{90}{motor} & \rotatebox{90}{bike} & mIoU & Avg.         \\ \midrule[0.5px]\midrule[0.5px]
\multirow{2}{*}{Data Comb.~\cite{ADVENT}}   & C               & 86.1          & 32.0              & 79.8              & 24.3          & 22.3           & 28.5          & 27.9           & 14.3          & 85.1          & 29.8             & 79.9         & 56.1            & 20.5           & 77.7         & \underline{34.4}           & 35.2         & \textbf{0.7}            & 18.2           & 13.1          & 40.3          & \multirow{2}{*}{41.2} \\
                              & I               & 92.8          & 23.4              & 60.9              & 25.8          & \underline{23.4}           & 24.1          & 8.6            & 32.2          & 77.5          & 26.8             & \underline{92.3}        & 48.0            & 41.0           & 74.4         & 48.4           & 17.7         & \textbf{0.0}       & 52.5           & 28.2          & 42.0          &                       \\ \midrule[0.5px]
\multirow{2}{*}{Individual~\cite{ADVENT}}   & C               & 88.8          & 23.8              & 81.5              & 27.7          & 27.3           & 31.7          & 33.2           & 22.9          & 83.1          & 27.0             & 76.4         & 58.5            & \underline{28.9}           & 84.3         & 30.0           & 36.8         & \underline{0.3}          & 27.7           & \underline{33.1}          & 43.3          & \multirow{2}{*}{43.5} \\
                              & I               & \underline{94.1}         & 24.4              & \textbf{66.1}             & \underline{31.3}          & 22.0           & 25.4          & 9.3            & 26.7          & \underline{80.0}          & \underline{31.4}             & \textbf{93.5}         & 48.7            & 43.8           & 71.4         & 49.4           & 28.5         & \textbf{0.0}           & 48.7           & \textbf{34.3}          & 43.6          &                       \\ \midrule[0.5px]
\multirow{2}{*}{CCL~\cite{CCL}}          & C               & \underline{90.3}          & \underline{34.0}              & \underline{82.5}              & 26.2          & 26.6           & 33.6          & \underline{35.4}           & 21.5          & 84.7          & \underline{39.8}             & \underline{81.1}         & 58.4            & 25.8           & \underline{84.5}         & 31.4           & 45.4         & 0.0            & 29.9           & 24.7          & 45.0          & \multirow{2}{*}{45.5} \\
                              & I               & \textbf{95.0}          & \underline{30.5}              & \underline{65.6}              & 29.4          & \underline{23.4}           & 29.2          & 12.0           & 37.8          & 77.3          & 31.3             & 91.9         & \textbf{52.4}            & 48.3           & 74.9        & 50.1           & 36.6         & \textbf{0.0}           & \underline{56.1}           & \underline{32.4}          & 46.0          &                       \\ \midrule[0.5px]
\multirow{2}{*}{ADAS~\cite{ADAS}}         & C               & -             & -                 & -                 & -             & -              & -             & -              & -             & -             & -                & -            & -               & -              & -            & -              & -            & -              & -              & -             & 45.8          & \multirow{2}{*}{46.1} \\
                              & I               & -             & -                 & -                 & -             & -              & -             & -              & -             & -             & -                & -            & -               & -              & -            & -              & -            & -              & -              & -             & 46.3          &                       \\ \midrule[0.5px]
\multirow{2}{*}{CoaST~\cite{CoaST}}        & C               & 81.7          & \textbf{38.3}             & 71.0              & \textbf{33.3}          & \underline{30.7}           & \underline{35.1}          & \textbf{38.2}           & \textbf{37.6}          & \textbf{86.4}        & \textbf{46.9}             & \textbf{81.9}        & \underline{63.4}           & 27.4          & \underline{84.5}        & 29.4           & \underline{45.6}        & \underline{0.3}           & \textbf{32.6}         & 31.3          & \underline{47.1}          & \multirow{2}{*}{\underline{48.2}} \\
                              & I               & 85.7          & \textbf{36.1}             & 65.1              & \textbf{33.2}          & \textbf{23.7}           & \textbf{32.8}          & \textbf{19.0}           & \underline{62.9}          & \textbf{82.5}          & 29.5             & 91.8         & \underline{52.1}           & \textbf{55.3}          & \textbf{83.4}         & \underline{62.9}           & \underline{46.1}         & \textbf{0.0}            & 55.5           & 18.5          & \textbf{49.3}         &                       \\ \midrule[0.5px]\midrule[0.5px]  \rowcolor{gray!15}
 & C               & \textbf{90.7}          & 31.0                 & \textbf{85.2}                  & \underline{31.0}              & \textbf{33.6}              & \textbf{35.2}             & 34.8             & \underline{32.0}              &  \underline{86.1}          &  38.6                & 79.6            & \textbf{65.0}                 & \textbf{32.4}              & \textbf{87.5}         & \textbf{44.9}            & \textbf{54.1}           & 0.0                & \underline{30.7}          & \textbf{51.4}            & \textbf{49.7}           &      \\  \rowcolor{gray!15}
\multirow{-2}{*}{OurDB (Ours)}  & I               & 92.8            & 24.3                 & 60.4                  & \textbf{33.2}             & 21.6              & \underline{32.6}          & \underline{14.6}              & \textbf{69.6}              & 76.0            & \textbf{33.6}               & 90.3             & 41.8                & \underline{52.2}               & \underline{79.7}           & \textbf{68.9}             & \textbf{59.3}         & \textbf{0.0}               & \textbf{58.2}             & 22.6            & \underline{49.0}             & \multirow{-2}{*}{\textbf{49.3}}                    \\ \midrule[1px]
\end{tabular}}
\label{tab:cls19}
\end{table*}

\textbf{19 classes benchmark.} In~\cref{tab:cls19}, we show the results on the 19 classes benchmark for the GTA5 $\rightarrow$ Cityscapes $+$ IDD combination. In terms of average mIoU, our method shows performance gaps of +5.8$\%$ compared to the Individual Model, and +1.1$\%$ compared to CoaST. In a detailed performance analysis, we observed that OurDB performs relatively well on classes that involve small objects. The results of all combinations on 19 classes benchmark are shown in~\cref{tab:cls19_all}. We achieve the highest performance in most of these experiments.

\subsection{Ablation Study}

\begin{table}[t]
\centering
\begin{minipage}[t]{.45\linewidth}
\centering
\caption{Comparison results on the 7 classes benchmark for all combinations. The best score is indicated in bold and the second best one is underlined.}
\label{tab:cls7_all}
\vspace{-3mm}
{\scriptsize
\begin{tabular}{ccc|c|ccc|c}
\midrule[1px]
\multicolumn{3}{c|}{Target}                                & \multirow{2}{*}{Method} & \multicolumn{3}{c|}{mIoU} & \multirow{2}{*}{Avg.} \\
C                 & I                 & M                 &                         & C      & I      & M      &                    \\ \midrule[0.5px] \midrule[0.5px]
\multirow{4}{*}{\checkmark} & \multirow{4}{*}{\checkmark} & \multirow{4}{*}{} & MTKT~\cite{MTKT}                    & 70.4       & 65.9       & -       & 68.2                      \\
                  &                   &                   & ADAS~\cite{ADAS}                    & \underline{75.4}       &  66.9      & -       & 71.2                       \\
                  &                   &                   & CoaST~\cite{CoaST}                   & 72.6       & \underline{70.0}       & -       & \underline{71.3}                      \\ \rowcolor{gray!20}
                 \cellcolor{white} & \cellcolor{white}                  & \cellcolor{white}                  & OurDB (Ours)            & \textbf{75.8}       & \textbf{71.2}       & -       & \textbf{73.5}                      \\ \midrule[0.5px]
\multirow{4}{*}{\checkmark} & \multirow{4}{*}{} & \multirow{4}{*}{\checkmark} & MTKT~\cite{MTKT}                    & 71.1       & -       & 70.8       & 70.9                      \\
                  &                   &                   & ADAS~\cite{ADAS}                    & \textbf{75.3}       & -       & \underline{72.6}       & \underline{73.9}                      \\
                  &                   &                   & CoaST~\cite{CoaST}                   & 72.3       & -       & 72.3       & 72.3                      \\ \rowcolor{gray!20}
                 \cellcolor{white} & \cellcolor{white}                  & \cellcolor{white}                  & OurDB (Ours)            & \underline{74.8}       & -       & \textbf{73.9}       & \textbf{74.4}                      \\ \midrule[0.5px]
\multirow{3}{*}{} & \multirow{3}{*}{\checkmark} & \multirow{3}{*}{\checkmark} & MTKT~\cite{MTKT}                    & -       & 65.9       &  70.7      & 68.3                      \\
                  &                   &                   & CoaST~\cite{CoaST}                   &  -      & \underline{68.7}       & \underline{72.4}       & \underline{70.6}                      \\ \rowcolor{gray!20}
                  \cellcolor{white} & \cellcolor{white}                  & \cellcolor{white}                  & OurDB (Ours)            & -       & \textbf{69.7}       & \textbf{74.1}       & \textbf{71.9}                      \\ \midrule[0.5px]
\multirow{4}{*}{\checkmark} & \multirow{4}{*}{\checkmark} & \multirow{4}{*}{\checkmark} & MTKT~\cite{MTKT}                    & 70.4       & 65.9       & 71.1       & 69.1                      \\
                  &                   &                   & ADAS~\cite{ADAS}                    & \underline{74.9}       & 66.7       & \underline{72.2}       & 71.3                      \\
                  &                   &                   & CoaST~\cite{CoaST}                   & 72.6       & \textbf{70.3}       & \underline{72.2}       & \underline{71.7}                      \\ \rowcolor{gray!20}
                  \cellcolor{white} & \cellcolor{white}                  & \cellcolor{white}                  & OurDB (Ours)            & \textbf{75.8}       & \underline{70.2}       & \textbf{74.5}       & \textbf{73.5}  \\ \midrule[1px]             
\end{tabular}}
\end{minipage}
\hspace{3mm}
\begin{minipage}[t]{.45\linewidth}
\centering
\caption{Comparison results on the 19 classes benchmark for all combinations. The best score is indicated in bold and the second best one is underlined.}
\label{tab:cls19_all}
\vspace{-3mm}
{\scriptsize
\renewcommand{\arraystretch}{0.94}
\begin{tabular}{ccc|c|ccc|c}
\midrule[1px]
\multicolumn{3}{c|}{Target}                                & \multirow{2}{*}{Method} & \multicolumn{3}{c|}{mIoU} & \multirow{2}{*}{Avg.} \\
C                 & I                 & M                 &                         & C      & I      & M      &                    \\ \midrule[0.5px] \midrule[0.5px]
\multirow{4}{*}{\checkmark} & \multirow{4}{*}{\checkmark} & \multirow{4}{*}{} & CCL~\cite{CCL}                    & 45.0       & 46.0       & -       & 45.5                      \\
                  &                   &                   & ADAS~\cite{ADAS}                    & 45.8       &  46.3      & -       & 46.1                       \\
                  &                   &                   & CoaST~\cite{CoaST}                   & \underline{47.1}       & \textbf{49.3}       & -       & \underline{48.2}                      \\ \rowcolor{gray!20}
                 \cellcolor{white} & \cellcolor{white}                  & \cellcolor{white}                  & OurDB (Ours)            &\textbf{ 49.7}       & \underline{49.0}       & -       & \textbf{49.3}                      \\ \midrule[0.5px]
\multirow{4}{*}{\checkmark} & \multirow{4}{*}{} & \multirow{4}{*}{\checkmark} & CCL~\cite{CCL}                    & 45.1      & -       & 48.8      & 47.0                     \\
                  &                   &                   & ADAS~\cite{ADAS}                    & 45.8       & -       & 49.2       & 47.5                      \\
                  &                   &                   & CoaST~\cite{CoaST}                   & \underline{47.9}       & -       & \underline{51.8}      & \underline{49.9}                      \\ \rowcolor{gray!20}
                 \cellcolor{white} & \cellcolor{white}                  & \cellcolor{white}                  & OurDB (Ours)            & \textbf{49.0}       & -       & \textbf{52.3}      & \textbf{50.6}                      \\ \midrule[0.5px]
\multirow{3}{*}{} & \multirow{3}{*}{\checkmark} & \multirow{3}{*}{\checkmark} & CCL~\cite{CCL}                    & -       & 44.5       &  46.4      & 45.5                      \\
&                   &                   & ADAS~\cite{ADAS}                   &  -      & 46.1       & 47.6       & 46.9                      \\
                  &                   &                   & CoaST~\cite{CoaST}                   &  -      & \textbf{49.5}       & \underline{51.6}       & \textbf{50.6}                      \\ \rowcolor{gray!20}
                  \cellcolor{white} & \cellcolor{white}                  & \cellcolor{white}                  & OurDB (Ours)            & -       & \underline{48.1}       & \textbf{52.5}       & \underline{50.3}                      \\ \midrule[0.5px]
\multirow{4}{*}{\checkmark} & \multirow{4}{*}{\checkmark} & \multirow{4}{*}{\checkmark} & CCL~\cite{CCL}                    & 46.7       & 47.0      & 49.9      & 47.9                     \\
                  &                   &                   & ADAS~\cite{ADAS}                    & 46.9       & 47.7       & 51.1       & 48.6                      \\
                  &                   &                   & CoaST~\cite{CoaST}                   & \underline{47.2}       & \underline{48.7}       & \underline{51.4}      & \underline{49.1}                      \\ \rowcolor{gray!20}
                  \cellcolor{white} & \cellcolor{white}                  & \cellcolor{white}                  & OurDB (Ours)            &  \textbf{49.6}    & \textbf{48.8}     & \textbf{52.9}   & \textbf{50.4}  \\ \midrule[1px]             
\end{tabular}}
\end{minipage}
\end{table}

\textbf{Impact of components.} We conduct ablation studies to investigate the impact of each component on our method. In~\cref{tab:ablation_all}, the first row presents the results for the baseline, which is devoid of any components proposed in this paper. To resolve the biased alignment instigated by aligning multiple target domains with only a single teacher model, we incorporate an ODS module into the baseline, the aftermath of which is depicted in the second row. We observe that the ODS module improves performance by $+$1.4$\%$. This proves that the ODS module effectively suppresses the biased alignment problem. It also indicates that the MTDA problem can be solved by changing the learning process instead of increasing the number of models. The third row is the result of additionally applying the AF-EMA. AF-EMA is designed to prevent the teacher model from forgetting knowledge of the previous target domain. We confirm that AF-EMA mechanism works effectively as intended, which is shown by the $+$0.8$\%$ gain. In addition, the enhancement of $+$1.3$\%$ in model performance through CGMix underscores the substantial advantages of leveraging contextual information in semantic segmentation tasks. As a result, we substantiate through this experiment that each of the components proposed in this paper imparts a positive impact on solving the MTDA problem.

\begin{table}[t]
\centering
\begin{minipage}[t]{.45\linewidth}
\centering
\caption{Ablation results of analyzing the impacts of our components on 7 classes benchmark for the GTA5 $\rightarrow$ Cityscapes + IDD combination.}
\label{tab:ablation_all}
\vspace{-3mm}
{\scriptsize	
\renewcommand{\arraystretch}{1.15}
\begin{tabular}{ccc|cc|c}
\midrule[1px]
ODS & AF-EMA & CGMix & C & I & Avg. \\ \midrule[0.5px]\midrule[0.5px] 
 &              &       & 72.8  & 67.12  & 70.0 \\ \rowcolor{gray!20}
 \checkmark &              &       & 74.1  & 68.7  & 71.4 \\
 \checkmark & \checkmark   &       & 74.5  & 69.9  & 72.2 \\ \rowcolor{gray!20}
 \checkmark & \checkmark   & \checkmark      & \textbf{75.8}  & \textbf{71.2}  & \textbf{73.5} \\
\midrule[1px]
\end{tabular}}
\label{tab:ab1}
\end{minipage}
\hspace{3mm}
\begin{minipage}[t]{.45\linewidth}
\centering
\caption{Comparison results based on different orders of target domains. We conduct experiments on 7 classes benchmark.}
\label{tab:ablation_order}
\vspace{-3mm}
{\scriptsize	
\begin{tabular}{c|c|c|c}
\midrule[1px]
 Order & Target & mIoU & Avg. \\ \midrule[0.5px] \midrule[0.5px] 
 \multirow{2}{*}{Cityscapes $\rightarrow$ IDD} & C & 75.8 & \multirow{2}{*}{73.5} \\
 & I & 71.2 & \\ \midrule[0.5px] 
 \multirow{2}{*}{IDD $\rightarrow$ Cityscapes} & C & 75.7 & \multirow{2}{*}{73.4} \\
 & I & 71.2 & \\
 \midrule[1px]
\end{tabular}}
\end{minipage}
\end{table}

\begin{figure*}[t]
    \centering
    \includegraphics[width=1\linewidth]{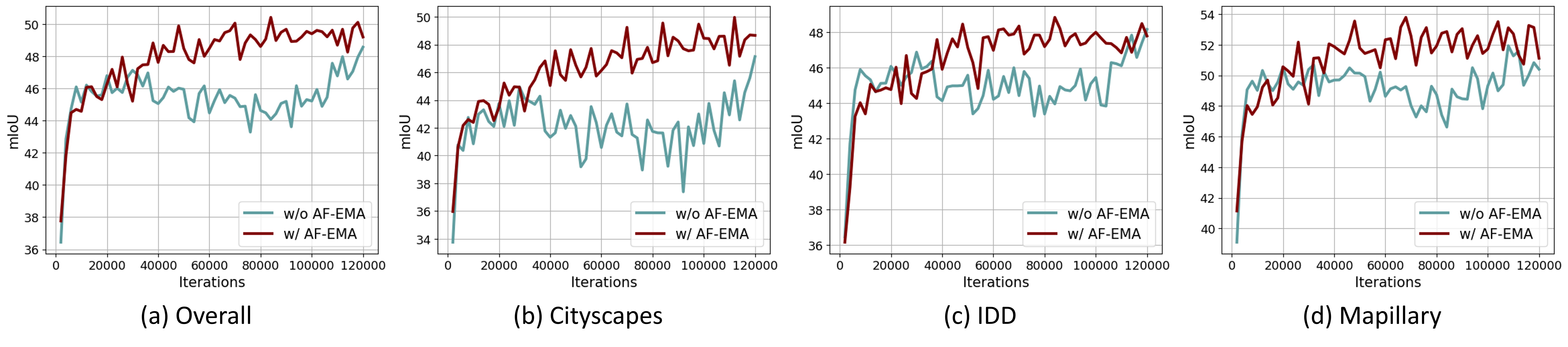}
    \vspace{-5mm}
    \caption{Performance comparison with and without AF-EMA on 19 classes benchmark for GTA5 $\rightarrow$ Cityscapes $+$ IDD $+$ Mapillary combination.}
    \label{fig:ablation_afema}
\end{figure*}

\textbf{Analysis of the ODS.} We conduct further experiments with the order of the target domains in the ODS module. As ODS cyclically selects one domain from multiple target domains, the order of the target domains can potentially influence the model's performance. We investigate the impact of the order of the target domains through two scenarios on 7 classes benchmark for GTA5 $\rightarrow$ Cityscapes $+$ IDD combination, i.e., Cityscapes $\rightarrow$ IDD and IDD $\rightarrow$ Cityscapes. As shown in~\cref{tab:ablation_order}, the performance difference between the two scenarios is a mere 0.1$\%$, indicating that the order of the target domains does not significantly impact the final performance of OurDB. More results are provided in the supplementary material.

\textbf{Analysis of the AF-EMA.} We visualize the evolution of performance over time with and without AF-EMA in~\cref{fig:ablation_afema}. These show the experimental results on 19 classes benchmark for GTA5 $\rightarrow$ Cityscapes $+$ IDD $+$ Mapillary combination. We observe that the performance of the model gradually deteriorates when AF-EMA is not applied. This stems from the problem of the teacher model forgetting the knowledge of the previous target domain. Although there is some improvement in performance during the latter part of training, it is challenging to narrow the performance gap with the model using AF-EMA. This demonstrates that the AF-EMA mechanism is essential to align multiple target domains through ODS.

\begin{figure*}[t]
    \centering
    \includegraphics[width=1\linewidth]{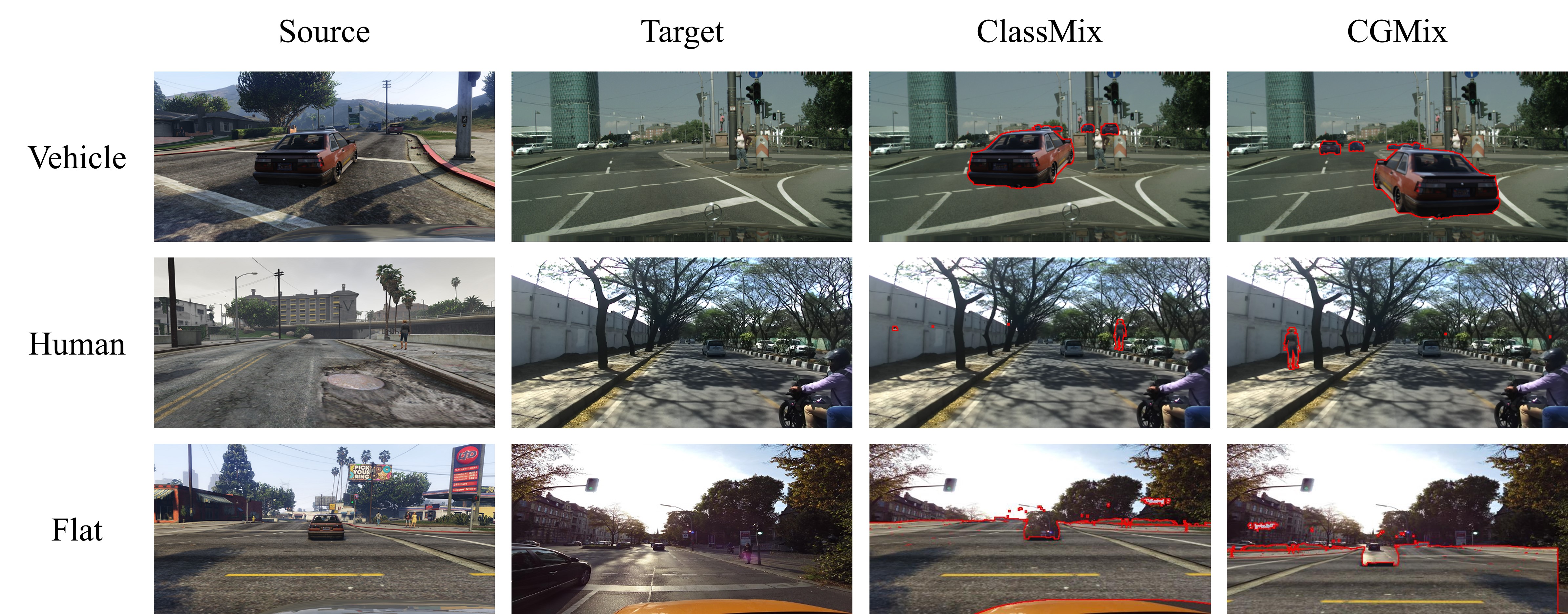}
    \vspace{-5mm}
    \caption{Qualitative comparison between ClassMix and CGMix on GTA5, Cityscapes, IDD, and Mapillary datasets. GTA5 is treated as the source domain, and the others are regarded as the target domains. The first row presents the results of bridges between GTA5 and Cityscapes, and the second and third rows depict the results between GTA5 and IDD, and GTA5 and Mapillary, respectively. The classes inserted into the target images are marked with a red outline.}
    \label{fig:qualitative_cgmix}
    \vspace{-2mm}
\end{figure*}

\textbf{Analysis of the CGMix.} Our method yields bridges with uncontaminated contextual information through CGMix. ~\cref{fig:qualitative_cgmix} presents a qualitative comparison between ClassMix~\cite{ClassMix} and CGMix on the GTA5~\cite{GTA5}, Cityscapes~\cite{Cityscapes}, IDD~\cite{IDD}, and Mapillary~\cite{Mapillary} datasets. The first and second rows are the results of the bridges generated by ClassMix and CGMix for the \textit{vehicle} and \textit{human} classes, respectively. In the source data, \textit{vehicle} and \textit{human} classes are primarily located on \textit{flat} class, yet these contextual relationships are not maintained in bridges built by ClassMix. For instance, some vehicles appear on the \textit{object} or \textit{sky} classes, and humans are situated within the \textit{nature} or \textit{construction} classes in the bridges of ClassMix. It suggests that the contextual information is corrupted in these bridges. The third row shows the results for the \textit{flat} class. Since more than half of the pixels in the source data belong to the \textit{flat} class, pasting the \textit{flat} directly onto the target data would occlude the existing classes. CGMix resolves these issues by adjusting the spatial location of each class based on the contextual relationships among the classes. In the bridges of CGMix, the \textit{vehicle} and \textit{human} classes are positioned on the \textit{flat}, and the \textit{flat} class is placed so as to minimally obscure other classes (e.g., \textit{construction} and \textit{nature}).

\section{Conclusions} In this paper, we investigated a method that can effectively address MTDA problems with only a single teacher architecture. We identify the limitation of the previous multiple teacher architectures and introduce three methods to overcome it. We propose ouroboric domain selector and anti-forgetting EMA to tackle the problem of biased alignment and forgetting the knowledge of the previous target domain. In addition, we enable the model to leverage contextual information tailored to diverse target contexts through context-guided class-wise mixup. We have demonstrated the effectiveness of our approach against the state-of-the-art methods through extensive experiments.

\bibliographystyle{splncs04}
\bibliography{main}
\end{document}